# What Can Robots Teach Us About Trust and Reliance? An interdisciplinary dialogue between Social Sciences and Social Robotics

Julien Wacquez[1], Elisabetta Zibetti[2], Joffrey Becker[1], Lorenzo Aloe[1,2], Fabio Amadio[3], Salvatore Anzalone[2], Lola Cañamero[1], Serena Ivaldi[3]

[1] *ETIS (UMR8051, CY Cergy Paris University, ENSEA, CNRS)*,

[2] *Laboratoire CHART (UR4004) Université Paris 8,* [3] *Inria, Université de Lorraine, CNRS*

*Abstract*— As robots find their way into more and more aspects of everyday life, questions around trust are becoming increasingly important. What does it mean to trust a robot? And how should we think about trust in relationships that involve both humans and non-human agents? While the field of Human–Robot Interaction (HRI) has made trust a central topic, the concept is often approached in fragmented ways. At the same time, established work in sociology, where trust has long been a key theme, is rarely brought into conversation with developments in robotics. This article argues that we need a more interdisciplinary approach. By drawing on insights from both social sciences and social robotics, we explore how trust is shaped, tested and made visible. Our goal is to open up a dialogue between disciplines and help build a more grounded and adaptable framework for understanding trust in the evolving world of human–robot interaction.

## I. INTRODUCTION

Over the past two decades, the notion of trust has come under intense scrutiny and collective reassessment in the field of social robotics. In a world where robots are increasingly integrated into daily human activities, the issue of trust became a fundamental prerequisite for enabling harmonious, effective and safe interactions.

In Human-Robot Interaction (HRI), trust is widely considered a multidimensional and interdisciplinary concept [1], [2], moving beyond the view of robots as mere automated systems, evaluated solely based on the reliability of their performance to the original context in which it was firstly developed (see [3] for a discussion on this in terms of HRI). However, in HRI, and more broadly in Human-Computer Interaction (HCI), the issue of trust has been approached in highly varied ways, drawing upon a wide diversity of theoretical frameworks. As a result, the HRI literature remains fragmented and lacks conceptual coherence, making it particularly challenging to model or measure trust in artefacts (see [4] for a comprehensive synthesis).

Yet, from a broader perspective, the work on trust makes little reference to existing work in sociology, a discipline with a long-standing and rich tradition of theorizing trust. Conversely, sociologists have paid limited attention to how trust is being redefined and operationalized within social robotics, despite the conceptual and empirical challenges it raises.

One of the main problems in trust research lies in the failure to acknowledge its subjective and **multidimensional nature**. As pointed out in [4], the proliferation of overlapping and sometimes incompatible models has led to conceptual ambiguity and low adoption in practice. Moreover, the rapid evolution of technologies such as AI, social robotics and automation renders domain-specific trust models quickly obsolete and calls for constant **epistemological revision** and technological/computational **innovations**.

This article sets out to explore the possibility of an interdisciplinary dialogue between social robotics and sociology. It asks whether and to what extent, the notion of trust might serve as a conceptual bridge through which these two fields —often grounded in divergent assumptions and methods— can meaningfully inform each other.

Far from undermining the concept, this theoretical instability reinforces the rationale for adopting an interdisciplinary approach, particularly in the analysis of human–robot interactions, which are inherently asymmetrical [5]. These interactions involve novel forms of engagement between humans and non-human agents and invite us to question the relevance of traditional notions such as "trust." It is precisely here that "failure" becomes analytically meaningful: they provide an heuristic lens through which the solidity and limits of trust as a conceptual tool could be explored.

By bringing together insights from HRI, social sciences, and robotics, we aim to better articulate the theoretical and practical challenges that trust poses in rapidly evolving socio-technical environments. Our contribution is situated within this perspective: we interrogate trust not as a given, but as a contextual and evolving construct, made visible and testable through experimentation, especially in moments of disruption, malfunction or breakdown of the robotic artefact.

## II. ATTEMPTS AT DEFINING TRUST IN SOCIAL ROBOTICS

The role of trust in HRI has been widely discussed in recent years. Various definitions [6] and conceptual models of trust have been proposed (see e.g., [1], [4] and [7] for meta-analysis) alongside diverse methodological approaches to assess trust dimensions in automation (e.g., [8]) and in HRI (e.g., [2], [9], [10]). However, despite these efforts, there still remains a lack of consensus on the definition of trust [11], both in human–human interactions [6] and in HRI [12].

In HRI, trust is often conceptualized as a perceptual, cognitive and affective attitude based on human evaluations of robot attributes such as appearance, performance, autonomy level and personality traits [7]. A prevalent definition frames trust as the willingness to accept vulnerability to the actions of another based on positive expectations [13]. According to this still prevalent view in HRI, trust can be defined as "the reliance by an agent that actions prejudicial to their well-being will not be undertaken by influential others" [1, p. 24] as well as "the attitude that an agent will help achieve an individual's goals in a situation characterized by uncertainty and vulnerability" [14, p. 54]. Moreover, several factors have been identified since the beginning of the research in automation as impacting trustworthiness (see [14] and [15] for an extensive review). Because trust also depends on results, substantial research has indicated that humans adjust their trust in automation to different performance aspects such as reliability, validity, predictability and dependability (e.g., failure occurrences). Finally, emerging perspectives advocate for viewing trust in HRI as a social and multifaceted construct, incorporating moral and ethical dimensions beyond mere technical competence and reliability (e.g., [2]). In line with this, although attributing moral qualities to contemporary robots may initially appear counterintuitive or contentious (e.g., [16]), such attributions can be understood through a phenomenological lens of research that questions the human propensity to perceive robots as capable of "moral agency" especially in the frame of different robot's failure [17].

Trust, understood as a dynamic process, in contrast to a dichotomous distinction between trust and no trust [11], it is built upon initial expectations and evolves with new experiences and learning [18]. It can be undermined by negatively perceived events (trust violations), which have a greater impact than positive ones due to their often visible consequences [14]. Although such events do not necessarily lead to a loss of trust, thanks to mechanisms of reinterpretation [19] trust repair strategies remain particularly challenging, especially in HRI [17], [20].

As technology becomes increasingly fine-tuned and versatile, this evolution invites us to re-question our conceptions of trust in robots and consider whether assistive robots, which support us in the accomplishment of specific but dissimilar daily tasks such as serving or giving care, can increase expectations of their reliability or even be viewed as a potential trusted agents. Yet, it is frequently assumed that the psychological mechanisms underpinning human trust in intelligent machines (i.e., autonomous vehicles, AI assistants, or social robots) mirror those of human interpersonal trust. However, this assumption may overlook critical distinctions.

## III. TRUST IN SOCIAL ROBOTICS: THE SOCIOLOGY PERSPECTIVE

In most research on trust within the field of social robotics, "human-robot interaction" tends to be approached *analogically* to "human-human interaction." The latter serves both as a conceptual model and as a normative reference point for understanding and designing the former. The aim of this analogy is often to foster in users a form of trust toward robots that *mirrors* interpersonal trust between humans.

Social theory supports the idea that it is difficult to generalise the notion of trust outside human interactions. Traditionally, sociology has been interested in the different ways in which trust manifests itself. Among these modalities, two will serve as a basis: interpersonal trust and trust in institutions [21].

Sociologists such as Niklas Luhmann suggest that trust involves a reflexive exercise about interactions and therefore the memory of previous interactions. Luhmann describes this dimension, which seems to be absent from work in HRI, as a necessary condition for achieving a kind of cognitive economy that enables the person using it to reduce the intrinsic complexity of a situation [22, pp. 26–28]. To be cognitively effective, this reduction must be based on situated learning and the generalisation of experience. It has the effect of creating a climate of familiarity [22, p. 33], established throughout life by socialisation, which makes it possible to anticipate. If individuals are "naturally" disposed to trust, it is because they have to manage risks, take risks, and consequently situate themselves in the interaction. This forms an operational level for trust, which lies at the level of interpersonal relations.

At another level, sociologists are interested in trust in systems such as institutions [21]. This level is not exclusive of the first, since it generally serves as a framework that enables individuals to organise their experiences both hermeneutically and in terms of their learning. Luhmann notes that interpersonal trust is rooted in individual experience and learning, but that systemic trust plays an important role in supporting the learning of trust, with social systems acting as frameworks for learning. These social systems can be seen as abstractions, albeit with very concrete implications from the individual's point of view (e.g. from a moral perspective). These abstractions are organised into highly sophisticated socio-technical systems, and take a variety of forms. The sociologist Georg Simmel has studied them, taking an interest, for example, in the establishment of loans and credit in banking systems; a technical system where interpersonal trust and systemic trust come together [23].

Yet trust is not necessarily confined to interpersonal relationships. It is closely linked to institutions. A vast sociological literature has examined how trust operates in relation to complex systems and institutions—such as governments, media, public services, healthcare systems, or science itself—particularly in the context of a *growing culture of suspicion* and a broader (political) *climate of mistrust*. These forms of systemic or institutional trust (or mistrust) are almost entirely absent from current research in social robotics (see [11]), even though they could be relevant, insofar as robots are designed, built, and deployed by institutions that may or may not be trusted, depending on the social and professional groups to which people belong (see e.g., [24]).

The predominant analogy with interpersonal trust is usually justified in two ways. First, it draws on the human tendency to implicitly anthropomorphize robots [25] while simultaneously denying their humanity [26]. This reasoning assumes that if individuals interact with robots in ways similar to how they engage with other humans, they may also be inclined to extend a form of "trust" to them. Second, the analogy is supported by the consideration that robots are no longer perceived merely as "tools," but increasingly as "teammates," "partners," or "agents" capable of performing increasingly complex tasks or exhibiting sophisticated behaviours (e.g., [27]). From a sociological perspective, however, an interaction between two humans cannot (at least, not yet) be equated with an "interaction" between a human and a robot. The fundamental reason why interpersonal trust cannot simply be transposed to human-robot interactions is that, due to their limitations, the latter does not reproduce "the structural conditions that make [trust] possible" [28, p. 48].

### III.A Reciprocity

The first of these conditions is that trust arises through **mutual engagement**. It is, by definition, a reciprocal commitment between the parties involved. When two individuals establish that they are in a trust relationship, a new normative and interpretive framework comes into effect for both: what they can or cannot do, must or must not do, and are or are not allowed to do, is entirely redefined by this new frame.

To speak of "trust in the robot," scholars in social robotics are necessarily led to adopt a reduced or impoverished conception of trust as a **unidirectional phenomenon**. [5] acknowledge this by favouring a "trust model" derived from management sciences, specifically developed by [13]:

> "trust between the trustor and the trustee *does not depend on* the trustee also trusting the trustor […] This asymmetry in trust is *an important prerequisite for transferring the interpersonal definition of trust to the definition of human trust in robots*, because it is not necessary (and currently not possible) for a robot to reciprocate trust in order for the trustor to trust the robot. […] If trust relations are asymmetrical, it is irrelevant whether robots are themselves able to trust and thus possess human-like capacities that would allow trust, e.g., intentionality" ([5, pp. 4–7], emphasis ours).

This is more than a terminological disagreement: it raises the question of whether the two disciplines are truly studying the same thing when they approach the topic of trust. We argue that this discrepancy is epistemological in nature and carries far-reaching implications for the understanding of trust as a social phenomenon.

### III.B Betrayal

The second structural condition of trust lies in the possibility that the other party might betray or deceive the trustor if they see significant personal gain in doing so. The vulnerability of the trusting person, therefore, does not stem from the mere risk of unmet expectations. Annette Baier draws a distinction between, on the one hand, simple ***disappointment***, that is, the failure to fulfil an expectation or hope, along with the ensuing let down, and on the other hand, ***betrayal*** or ***deception*** [29]. Betrayal necessarily involves both the pursuit of self-interest and the intention to harm. In this regard, the robot can certainly disappoint, because of a malfunction or because its capabilities have been overestimated. But it cannot, strictly speaking, betray.

### III.C Asymmetry (as a submission to the other's point of view)

The asymmetry inherent in any relationship of trust, then, is not due to its supposed unidirectionality, as proposed by [5], but to the fact that the trustor entrusts the trustee with discretionary power, thereby effectively submitting to their will and judgment. This is a point explicitly made by Lars Hertzberg when comparing what he calls the grammar of trust to the grammar of reliance:

> "The grammar of reliance fits a situation in which I take up an attitude towards a person because I believe certain things about him; a situation involving trust, on the other hand, is one in which I may come to believe certain things because I have an attitude towards a person. When I trust someone, it is *him* I trust; I do not trust certain things *about* him. If I trust someone there cannot be certain respects in which I distrust him. Distrusting him would mean that I myself retained the ultimate judgment concerning the respects in which he was to be trusted; but then I would not really have placed my trust in *him*.
>
> In relying on someone <...> I look down at him from above. I exercise my command of the world. I remain the judge of his actions. In trusting someone I look up from below. I learn from others what the world is about. I let him be the judge of my actions." ([30, pp. 314–315], emphasis in the original).

If the trustee's perspective is deemed irrelevant, as [5] suggests, then the asymmetry is of a different nature, and it is likely that we are not dealing with a situation of trust, but rather one of reliance.

In other words, if we ignore the perspective of the person/agent we "trust" (i.e., the robot's intention, its understanding of the situation or its degree of autonomy), then it is not truly a case of trust. Rather, we are merely considering the robot as a reliable tool, which falls under reliance, and not under trust.

### III.D Claims

By submitting to the judgment of the other, the trustor thus makes themselves vulnerable, in the sense that the trusted person has always the possibility to betray them, to take advantage of them. But this vulnerability allows the trustor to shape the relationship and, in doing so, to raise claims about how the other party ought to behave:

> "Being the recipient of trust creates specific obligations—chief among them, the duty to honour that trust by behaving in a certain way, and the duty not to take advantage of the vulnerability of the trust-giver […]. For the trustee, fulfilling these obligations is a task that opens the door to failure, clumsiness, lack of tact, missteps, duplicity, or the temptation to abuse the trust placed in them." [21, p. 230]

This means that vulnerability does not rest solely with the trustor; it is shared between both parties. Indeed, if the trustee fails to fulfil their obligations, they become subject to normative expectations and the judgment of others.

Conversely, the person who is trusted can also make demands regarding how the trustor ought to behave. As [21] points out, the trustor is expected to suspend their own judgment, refrain from constant interference, relinquish excessive monitoring and control, and avoid seeking unnecessary reassurance.

If sociologists understand trust as a mutual engagement, it is because it transforms the behaviour of both the trustor and the trustee. It implies obligations on both sides, as well as the possibility of making normative claims. But it would be incongruous for a human to say to a robot: "*Treat me the way I deserve to be treated!*"

**III.E Temporary conclusion**

Because human-robot interaction fulfils none of the structural conditions that make interpersonal trust possible—reciprocity, the possibility of betrayal, submission to the other's point of view, and the ability to make normative claims—we may legitimately question whether trust is genuinely at stake in such interactions.

A sociologist could very well conclude that researchers in social robotics are committing a category error: they believe they are working on trust, when in fact they are dealing with something entirely different. What exactly? That remains to be determined.

However, as we have already suggested, there are multiple forms of trust, and not all of them operate at the interpersonal level.

In order to find a middle ground and make dialogue between sociology and social robotics possible, we must therefore move beyond the narrow case of interpersonal trust and consider trust as a way for people to **engage in any situations**, as a **pre-reflexive** response to them.

**IV. TOWARDS A COMMON GROUND: TRUST AND PRACTICAL ENGAGEMENT**

If we adopt a broader view of trust, that is not merely as a reciprocal moral commitment, but as a form of practical engagement with the world, a pre-reflective mode of being toward an environment or artefacts, insofar as it is more cognitively economic way to act in the world, including in relation to technologies, then it becomes possible to explore common ground between the social sciences and social robotics. It is from this broader perspective that we propose to reconceptualise the notion of trust in HRI. Rather than attempting to replicate the structures of interpersonal trust, the focus shifts to understanding how humans engage with robots in concrete situations within rich and dynamic ecological contexts, and how robots, in turn, can adapt to these interactions.

This shift in perspective is particularly important when studying phenomena such as ***trust repair*** following an ***error*** or malfunction. While a robot cannot "betray" in the moral sense, it can nonetheless violate implicit expectations, leading the human counterpart to experience discomfort, a sense of rupture, or even disengagement [17], [31]. Indeed, several studies have shown that even in the absence of intentionality, humans often project normative expectations onto robotic systems and their breach can lead to similar emotional and behavioural responses as in human-human trust violations, even if certain nuances exist in participants' perceptions regarding trust violations based on competence and integrity, which could reflect ontological differences between humans and machines [17].

However, while providing great advances, current approaches in social robotics often rely on highly controlled and scripted scenarios, often through online video (e.g., [17]), which limit the ecological validity of findings and constrain our understanding of how trust is built, eroded, and potentially repaired in naturalistic settings.

To move beyond these limitations, but also to have a better understanding of the difference between H-H and H-R trust, it seems essential to endow robots with ***adaptive capabilities***: the ability to detect a breakdown in trust, to select and enact an appropriate trust repair strategy (e.g., offering an apology, providing an explanation, or adjusting behaviour) and to modulate this response based on real dynamic social context (e.g. a cafeteria, a restaurant, a supermarket, etc.) and potentially also based on users characteristics. Such adaptation presupposes computational models of trust that are sensitive not only to performance outcomes but also to relational human nonverbal cues (elements shown to be fundamental in interpersonal trust dynamics).

This approach should open a promising interdisciplinary research agenda between social sciences and robotics, wherein trust is not reduced to a binary state, but understood as a dynamic, context-sensitive process, shaped by mutual adaptation and embedded in real-world interactional ecologies.

**V. MULTIMODAL TRUST MODELLING AND REPAIR**

Trust is a critical factor in HRI in many domains, from healthcare and industry to domestic and education. However, trust in a robot is dynamic and could be easily undermined when the robot makes mistakes or violates expectations [32]. A trust violation is any event where a robot's actions reduce the human's perception of the robot's trustworthiness or reliability. In HRI, trust violations can be categorized based on which facet of trust is broken [32]:
- Ability (the robot fails to perform as expected)
- Integrity (the robot breaks rules or promises)

5- Benevolence (the robot appears to act with ill intent)

Once trust is damaged, the robot (seen as the trustee) must take some action aimed to repair the trust after the violation.

Another important concept is trust calibration: ensuring the human's trust level matches the robot's true capabilities, neither too low (disuse) nor too high (misuse). Indeed, calibrating trust is often more desirable than simply maximizing it [17].

Recent investigations have offered useful perspectives on the effectiveness of different trust repair strategies such as apologies, promises, explanations and denials. However, a key limitation is that most of these studies have taken place in tightly controlled, scripted settings. For instance, [33] predefine not only when a trust violation occurs (e.g., a robot deliberately makes a mistake at a specific trial), but also what kind of trust repair strategy is employed (e.g., a fixed verbal message delivered immediately afterward). While this approach makes it easier to study human reactions to particular repair strategies, it doesn't reflect the complexity and unpredictability found in real-world interactions between humans and robots.

A promising idea is the one presented by [34] where they implement error-learning and adaptation as a form of trust repair: by quickly learning from a mistake and demonstrating improved performance, the robot shows competence and commitment to not repair the error. This integrated approach combining a social gesture (an apology) with a technical response (corrective action), offers a more robust framework for restoring trust. Users, in turn, were more inclined to re-establish trust in the robot when its apology was accompanied by tangible behavioural improvements.

To achieve effective trust repair mechanisms "in the wild", robots must autonomously recognize when trust has been damaged, select an appropriate repair strategy, deploy it with the right timing [31], potentially taking into account both contextual factors and human-specific traits. This requires moving beyond static and hand-crafted responses toward adaptive and real-time systems. In order to achieve this, it is essential to endow robots with functional computational models of human-robot trust. As identified by [35], some of the more well-established approaches are Performance-centric algebraic, Time-series, Markov Decision Process (MDP), Gaussian-based, and Dynamic Bayesian Network (DBN)-based models.

A thorough analysis of these models is beyond the scope of the current paper; we refer interested readers to the original survey [35]. However, regardless of their particular characteristics, all the previously mentioned types of models seem to share important common weaknesses:

They primarily rely on performance-related metrics, sometimes overlooking psychological and behavioural factors that play a crucial role in building human trust in robots. Model parameters are typically learned from human trust feedback data (often collected through questionnaires), while other types of implicit measure (reaction time, physical behavioural responses or psychophysiological signals) are rarely considered. Therefore, providing robots with the possibility to receive and interpret a broader range of human behavioural cues seems fundamental.

More recently there have been promising attempts to incorporate social signals and behavioural cues, key features include facial expressions, eye gaze, proxemics and vocal tone. These characteristics are important markers for trust and computer vision and audio processing techniques can be used to analyse this kind of cues, using machine learning classifiers to map them to trust levels [36], [37]. In particular, using a Random Forest classifier on facial features, 84% accuracy has been achieved in distinguishing trust vs. mistrust, highlighting the predictive power of facial signals [36]. It is worth noting that most of these cues must be analysed taking into account individual differences and context. For example, some users might mask their discomfort (deliberately smiling despite low trust), while others show micro-expressions of doubt. Lighting and camera angle can also affect automated detection reliability. Nonetheless, facial cues have to be considered into affective states related to trust. Exploiting them, it could be possible to adapt their communication. If a user seems distrustful, the robot might increase transparency or provide more frequent updates (proactive repair), if the user seems over-trusting, the robot might insert subtle warnings or ask for confirmation for critical steps. It would be useful to look at physiological and behavioural indicators of trust (e.g. heart rate changes, facial expressions and response delays) to infer when the user is not trusting the robot anymore. Assessing the trust dynamic is not trivial, and this is all the more true, given that trust is very rarely assessed behaviourally in real time [38], but rather reconstructed retrospectively via subjective self-reports [39]. Thus it should be done in real time and through robust metrics.

Indeed, exploiting modern vision algorithms, inertial measurement units (IMUs), microphones and social signal processing is possible to assess trust in real time [37], [40], [41].

Another important point is that also physiological signals are a powerful way to infer trust level, they provide an inside look at the user's internal state, often reflecting stress or arousal that might not be fully visible through outward behaviour. A major advantage is that physiology is hard to consciously control, so these indicators could be more reliable in situations where one would want to fake his emotional state, for example, keeping a poker face.

Main physiological metrics that can be used are:
- Heart rate (HR) and Heart rate variability (HRV) [37];
- Electro-dermal activity (EDA) [36];
- Electroencephalography (EEG) [42].

Obviously, these signals are context-dependant as well and their accuracy as to be improved (as can be seen in the cited papers, where the achieved accuracy is 68% in the case of HR and HRV experiments and 69% in the case of experiments in which EDA was used as a trust indicator), but they provide a piece of the puzzle in detecting user distrust. However, trust is a multifaceted human state, and no single signal is perfectly reliable. For this reason we would like to emphasize the importance of a multimodal approach, i.e. combining multiple sensors and cues to improve detection accuracy and robustness. Some promising results in this direction have been already achieved, proving that multimodal models achieve higher accuracy in trust detection than any single-modality level [43].

Another question to be addressed is: should a robot explicitly ask for feedback about trust? Sometimes clear (verbal) communication might be the best, but it can also put users on the spot. The best way would be to design specific algorithms for robots to assess trust (like observing if the human is double-checking what the robot is doing) and adapt to the user and the context. Recent advancements in Visual Language Models (VLMs) could constitute a potential way to make robots able to process both verbal and non-verbal cues and respond with the same channel, establishing an intuitive communication [44]. These skills could play a key role in achieving reactive and user-specific trust repair strategies, employing both verbal communication and adapting motion planning and control to better cope with the perceived level of human trust. However, it is an open question whether this adaptation mechanism is really necessary or not: in many application scenarios, robots could exhibit a lack of social intelligence and still be accepted and adopted.

Finally, it is also necessary to gain a better understanding of the underlying representations used by human partners in the course of the interaction itself, and to see to what extent their commitment and engagement are based on trust. By drawing on qualitative methods from sociology, anthropology, and psychology, in particular observation and interview techniques, it is possible to document in detail the description that users give of the interaction, and to estimate their trust. By focusing on the ways in which their practical engagement manifests itself, observation and interviews have two main benefits. The first is to better understand what status the robot occupies in the course of the interaction. The second is to gain a better understanding of the role of institutions (and consequently of the users' learning) in qualifying their relationship with the robot.

The articulation of these two dimensions (*quant* and *qual*) in the investigation of trust in human-robot interactions offer a particularly fruitful complementary framework for a mixed method [45]. It allows us to embrace different aspects characterising the complexity of a phenomenon such as trust by studying its many manifestations.

## VI. CONCLUSION

In this paper, we study the concept of trust in human robot interactions from a multidisciplinary perspective. After recalling the main theoretical foundations of trust in social robotics and sociology, we highlight the potential of multimodal approaches for modelling and repairing trust during human-robot interactions. We conclude with open questions related to the implementation of trust in social robots, such as how trust can be measured, and whether robots should explicitly seek and monitor the trust that humans exhibit during their interactions.

## ACKNOWLEDGMENTS

The authors and this work were funded by ANR through project PEPR O2R AS3 (ANR-22-EXOD-007).